\documentclass[lettersize,journal]{IEEEtran}
\usepackage{amsmath,amsfonts}
\usepackage{array}
\usepackage{textcomp}
\usepackage{subfigure}
\usepackage{stfloats}
\usepackage{url}
\usepackage{verbatim}
\usepackage{graphicx}
\usepackage{cite}
\usepackage{multicol}
\usepackage{multirow}
\usepackage{booktabs}
\usepackage{setspace}
\usepackage{tabularx}
\usepackage{makecell}
\usepackage{hyperref}
\usepackage{tabularx}
\usepackage[ruled, linesnumbered]{algorithm2e}
\usepackage[rgb]{xcolor}
\definecolor{deepgreen}{rgb}{0,0.5,0}
\begin{document}

\title{Hierarchical and Decoupled BEV Perception Learning Framework for Autonomous Driving}

\author{Yuqi Dai, Jian Sun, Shengbo Eben Li, Qing Xu, Jianqiang Wang, Lei He*, Keqiang Li

\thanks{
This study is supported by National Natural Science Foundation of China, Science Fund for Creative Research Groups (Grant No.52221005)

 Yuqi Dai, Jian Sun, Shengbo Eben Li, Qing Xu,  Jianqiang Wang, Lei He and Keqiang Li are with the School of Vehicle and Mobility, Tsinghua University, Beijing, China, and also with the State Key Laboratory of Intelligent Green Vehicle and Mobility, Tsinghua University, Beijing, China (e-mail: daiyuqi@mail.tsinghua.edu.cn; qingxu@tsinghua.edu.cn; lishbo@tsinghua.edu.cn; helei2023@tsinghua.edu.cn; likq@tsinghua.edu.cn)
\textit{(Corresponding author: Lei He.)}

Jian Sun is with the Department of Computer and Information Science, University of Macau, Macau, China, (e-mail: sjian1015@163.com)

}
}



\maketitle

\begin{abstract}

Perception is essential for autonomous driving system. Recent approaches based on Bird's-eye-view (BEV) and deep learning have made significant progress. However, there exists challenging issues including lengthy development cycles, poor reusability, and complex sensor setups in perception algorithm development process. To tackle the above challenges, this paper proposes a novel hierarchical BEV perception paradigm, aiming to provide a library of fundamental perception modules and user-friendly graphical interface, enabling swift construction of customized models. We conduct the Pretrain-Finetune strategy to effectively utilize large scale public datasets and streamline development processes. Moreover, we present a Multi-Module Learning (MML) approach, enhancing performance through synergistic and iterative training of multiple models. Extensive experimental results on the Nuscenes dataset demonstrate that our approach renders significant improvement over the traditional training scheme.

\end{abstract}



\section{Introduction}

\IEEEPARstart{A}{utonomous} driving, alternatively termed self-driving, is facilitated by the integration of sensor computing devices, information communication, automatic control, and artificial intelligence, transforming vehicles into entities capable of self-navigation \cite{yusuf2024vehicle}. The environmental perception system acts as the conduit for intelligent vehicles to assimilate external information. It is tasked with the collection, processing, and analysis of data pertaining to the vehicle's surroundings, thereby laying the groundwork and serving as a prerequisite for autonomous driving. As a pivotal element of intelligent driving technology, the perception system supplies essential input data for subsequent modules such as localization, prediction, decision-making, planning, and control \cite{han2023collaborative}. The perception system must be able to accurately identify a range of elements essential for navigation, including pedestrians, cyclists, vehicles, lights, as well as static road features such as road surfaces, lane markings and traffic signs. To ensure the safety and intelligence of autonomous vehicles, the algorithms underpinning the perception system must capable of reliably detecting the vehicle's dynamic surroundings from sensory data \cite{singh2023surround,gao2024roadside,zhao2024localization} and maintain remarkable performance.

At present, the majority of perception algorithms ready for on-vehicle implementation are predicated upon deep neural networks, which require significantly huge amounts of data resources and adequate computational resources for training. Currently the development process of perception algorithms faces several challenges: (1) The development cycles are often lengthy due to the need for extensive data collection, annotation, and format adaptation \cite{meng2023configuration}. As intelligent driving system platforms become increasingly diverse and the pace of feature updates accelerates, developers are prone to high development costs and repetitive procedures \cite{ma2023autors}. (2) Poor reusability of algorithms across different platforms poses a significant hurdle. The complexity involved in sensor setups adds to the difficulty of developing perception algorithms. Integrating and synchronizing sensors to ensure accurate and reliable data collection requires sophisticated calibration and coordination techniques.  Changes in sensor types, quantities, and layouts would necessitate brand new model adaptation and training. Moreover, the absence of a systematic approach to leveraging common functional modules during deployment has complicated the optimization of perception systems for mass production, hindering efficient adaptation to task-specific requirements. (3) A significant challenge in this domain is the generalizability of these algorithms across diverse and dynamic environments \cite{zhang2024feature}. Models that excel in one context may falter in another, emphasizing the necessity for adaptable and efficient perception systems. Mainstream solutions often lack functional software, fail to extract common algorithmic modules, and exhibit a low degree of component reuse. This lack of standardization necessitates substantial modifications and optimizations for each specific task, which has resulted in increased redundancy in engineering efforts, escalating development costs.

\begin{figure*}[t]
  \centering
  \includegraphics[width=\textwidth]{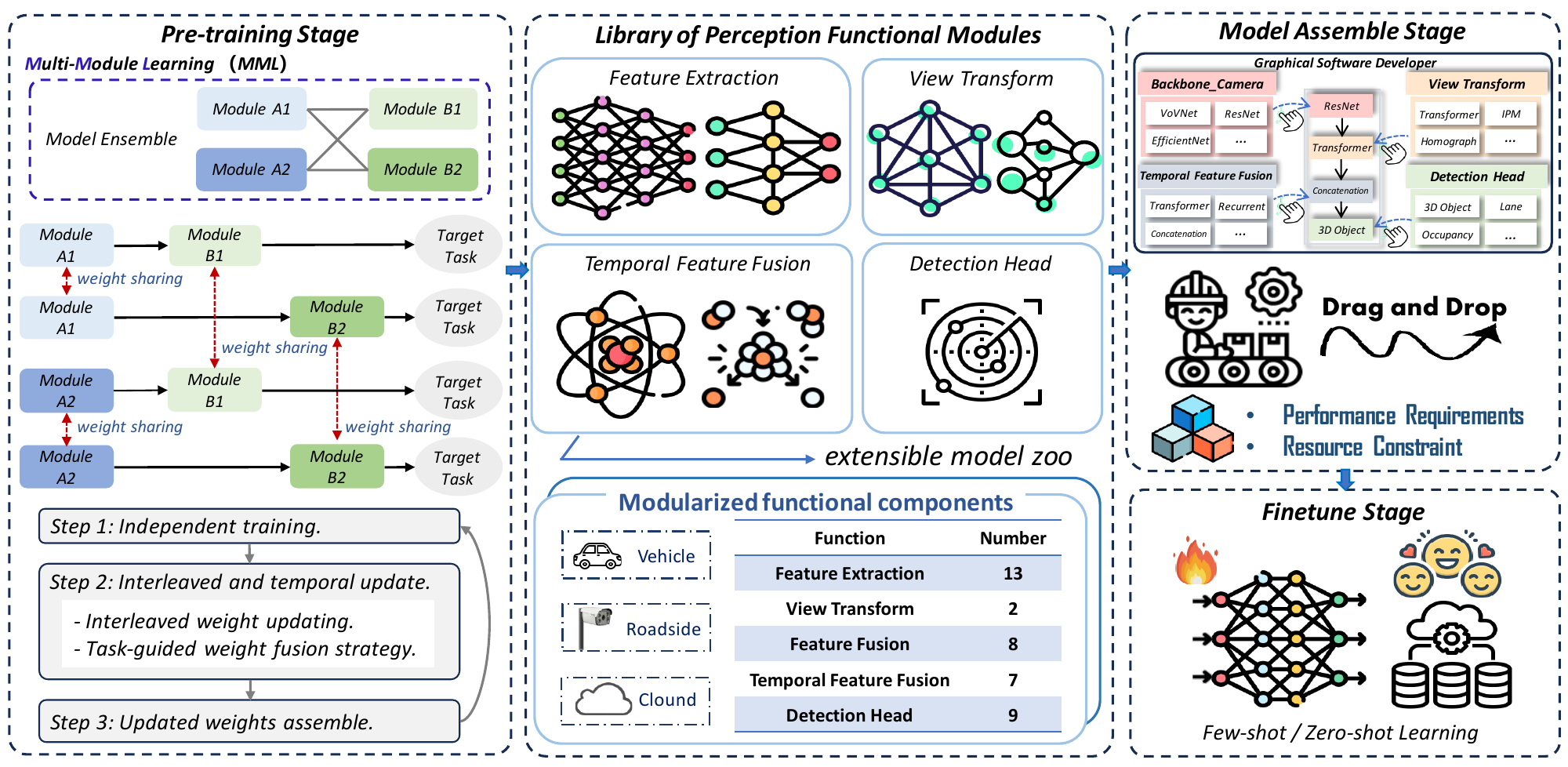}
  \caption{An overview of our hierarchical and decoupled BEV perception scheme: i) A perception model library is formed based on multi-module joint training. ii) Perception algorithm models are constructed through drag-and-drop operations using a graphical user interface. iii) Model fine-tuning is performed based on custom data.}
  \label{fig:hps}
\end{figure*}

In response to these challenges, this paper introduces a hierarchical and decoupled perception scheme, as shown in Fig.\ref{fig:hps}, in which a comprehensive framework is designed to streamline the development of perception algorithms for autonomous vehicles. By organizing fundamental algorithmic components into a functional module library, the proposed perception scheme empowers automotive developers to construct and tailor perception models that meet specific operational requirements with greater efficiency. Our approach to modularization categorizes perception models based on their functional roles, with each module providing a selection of network structures. This design not only facilitates alignment with the computational capabilities of various platforms but also simplifies the adaptation to custom datasets, thereby enhancing development efficiency. The high reusability of these standardized functional modules significantly reduces repetitive engineering efforts, leading to a condensed development timeline. Recognizing the rapid evolution of intelligent vehicle platforms and the demand for swift feature updates, our modular perception system is inherently extensible. It allows developers to introduce new modules or refine existing ones with ease, ensuring the system's evolution in response to emerging needs and technological advancements.

Furthermore, to enhance development efficiency, we propose a paradigm that integrates pre-training and fine-tuning. By leveraging extensive open-source datasets, we establish a diverse repository of well-performing models. These pre-trained models provide a solid foundation, requiring less training data and fewer iterations during the fine-tuning phase. This paradigm enables users to adapt functionalities and adjust parameters to suit custom datasets by means of a graphical interface, achieving optimal performance in real-world scenarios through techniques such as transfer learning and domain adaptation. 

Specifically, the pre-training regimen involves an exhaustive training of each potential module combination, ensuring that every perceptual module is endowed with compatible weights for both upstream and downstream functional counterparts. Once the perception functional module library is established, we introduce a novel Multi-Module Learning (MML) paradigm in the pre-training stage. Designed for the hierarchical and decoupled perception system, MML improves overall training efficiency and diverse model architectures can be optimized simultaneously to obtain performance improvements. MML framework has has demonstrated effectiveness in our empirical study.

In summary, this paper presents a transformative approach to the development of autonomous vehicle perception systems, offering a more efficient, adaptable, and scalable solution that. Our main contributions can be summarized as follows:

\begin{itemize}
  \item This paper introduces a hierarchical perception system, providing a library of fundamental components and graphical interface to streamline the development process. Users can effortlessly build their own perception models through drag-and-drop operations.
  \item The Pretrain-Finetune paradigm is employed to facilitate rapid deployment and enhance the generalizability of perception algorithm. This decoupled paradigm supports the standalone development of fundamental modules and promotes the highly flexible reutilization of components.
  \item We propose an innovative Multi-Module Learning (MML) framework, which has shown that coordinated training of multiple modules can improve individual task performance and enhance model robustness.
  \item Extensive experiments on 3D object detection demonstrate the effectiveness of the proposed BEV perception scheme.
\end{itemize}

\section{Related work}
In order to provide the foundational context for this study. This section synthesizes prior research on perception technologies, multi-task learning and joint training approaches within the domain of autonomous driving.

\subsection{Perception in Autonomous Driving}
Perception is integral to autonomous driving, serving as the system's sensory interface to the environment. Accurate perception is essential for making informed decisions about vehicle motion \cite{kuo2020autonomous}. Perception systems typically incorporate a variety of sensors to capture comprehensive data about the vehicle's surroundings. Cameras provide high-resolution imagery, LiDAR sensors offer precise distance measurements, and radars contribute with their ability to penetrate through various weather conditions \cite{shi2021survey, liu2022sensor}. The generation of BEV features has garnered substantial interest due to its holistic representation of the driving scene. In a BEV perception system, data from all sensors are processed within the BEV space, allowing for earlier fusion \cite{zhang2024multisensor}. Objects observed from BEV perspective are not subject to the scale inconsistencies and occlusion issues prevalent in image-based views. The BEV-based approach significantly enhances system capabilities, yielding more stable and accurate perception outcomes \cite{li2023delving,li2024fast}. Moreover, the unification of disparate viewpoints under the BEV framework substantially simplifies subsequent planning and control operations. Mainstream planning and control algorithms integrate the information and transform it into a vehicle-centric coordinate system, thereby aligning the output of BEV perception directly with the standard input required for planning and control tasks. Within the BEV space, both perception and prediction are conducted in the same spatial context, enabling end-to-end optimization through neural networks \cite{chib2023recent,chen2023end,hu2023planning}.

Early approaches to BEV feature generation relied on ConvNets and inverse perspective mapping (IPM) to transform perspective view features into a top-down representation \cite{zhou2022crossview}. Recently, the advent of Transformer architectures has marked a significant leap, showcasing their versatility and effectiveness in the realm of autonomous driving \cite{yang2021projecting, chen2022persformer, bevformer, beverse}. The key advantages of transformers are their capacity to integrate information across large receptive fields for comprehensive scene comprehension, a trait that sets them apart from CNNs. 




\subsection{Multi-task Learning and Joint Training.}
Multi-task architectures in deep learning typically employ a scheme known as hard parameter sharing. This involves utilizing a unified encoder that is shared across various tasks, complemented by distinct decoders tailored to each specific task \cite{zhang2018overview}. A notable example is MultiNet \cite{teichmann2018multinet}, which is a comprehensive feedforward architecture that consolidates three critical vision tasks: classification, detection, and semantic segmentation. The design is anchored on a common encoder that branches out into three specialized pathways, each equipped with multiple convolutional layers designed to cater to a specific task. For instance, a model trained for object detection can also learn to predict vehicle trajectories or classify road surfaces in a unified framework \cite{kim2020multi}.

\begin{figure}[!t]
	  \centering
		\subfigure[MTL]{
			\includegraphics[width=0.45\textwidth]{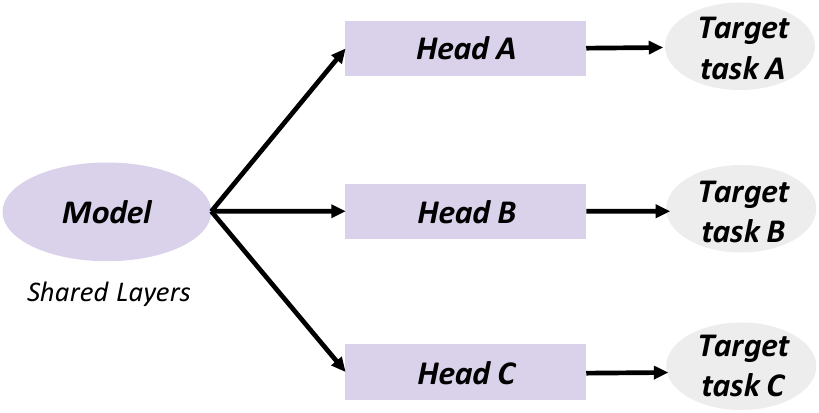}
		}
		\subfigure[MML]{
			\includegraphics[width=0.45\textwidth]{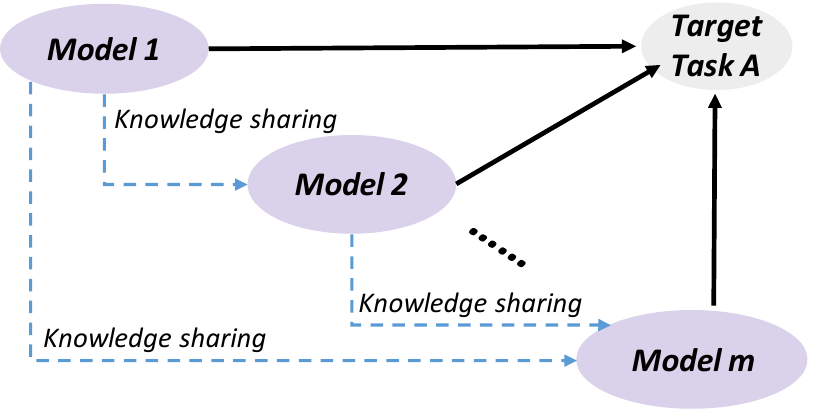}
		}
         \caption{Training framework overview for Multi-Task Learning and Multi-Module Learning.}
\label{fig:MTLvsMML}
\end{figure}

Multi-task learning has emerged as an effective strategy for training autonomous driving perception systems, allowing for the simultaneous optimization of multiple related tasks \cite{guo2023research}, as illustrated in Fig.\ref{fig:MTLvsMML}. This approach can lead to improved performance by leveraging shared representations and knowledge across multiple tasks \cite{ruder2017overview,shi2024collaborative}. DLT-Net \cite{qian2019dlt} integrates a context tensor within a multi-task learning framework to dynamically share spatial features across sub-tasks, while also employing a shared encoder-decoder structure to effectively model temporal dynamics and enhance feature representation for autonomous driving applications. YOLOP \cite{wu2022yolop} further exemplify the efficiency of multi-task learning by simultaneously addressing traffic object identification, drivable area detection, and lane line recognition within a single unified framework. Xia et al. \cite{xia2024henet} proposes HENet, a cutting-edge framework for multi-task 3D perception that uses a hybrid encoding approach and attention-based feature integration to enhance precision in autonomous driving scenarios. It adeptly manages task conflicts through specialized BEV feature encoding, achieving top-tier results on the nuScenes benchmark. These models highlight the potential of multi-task architectures to handle complex, multi-faceted problems, enhancing model generalization and streamlining computational resources. 

\begin{figure*}[t]
  \centering
  \includegraphics[width=0.9\textwidth]{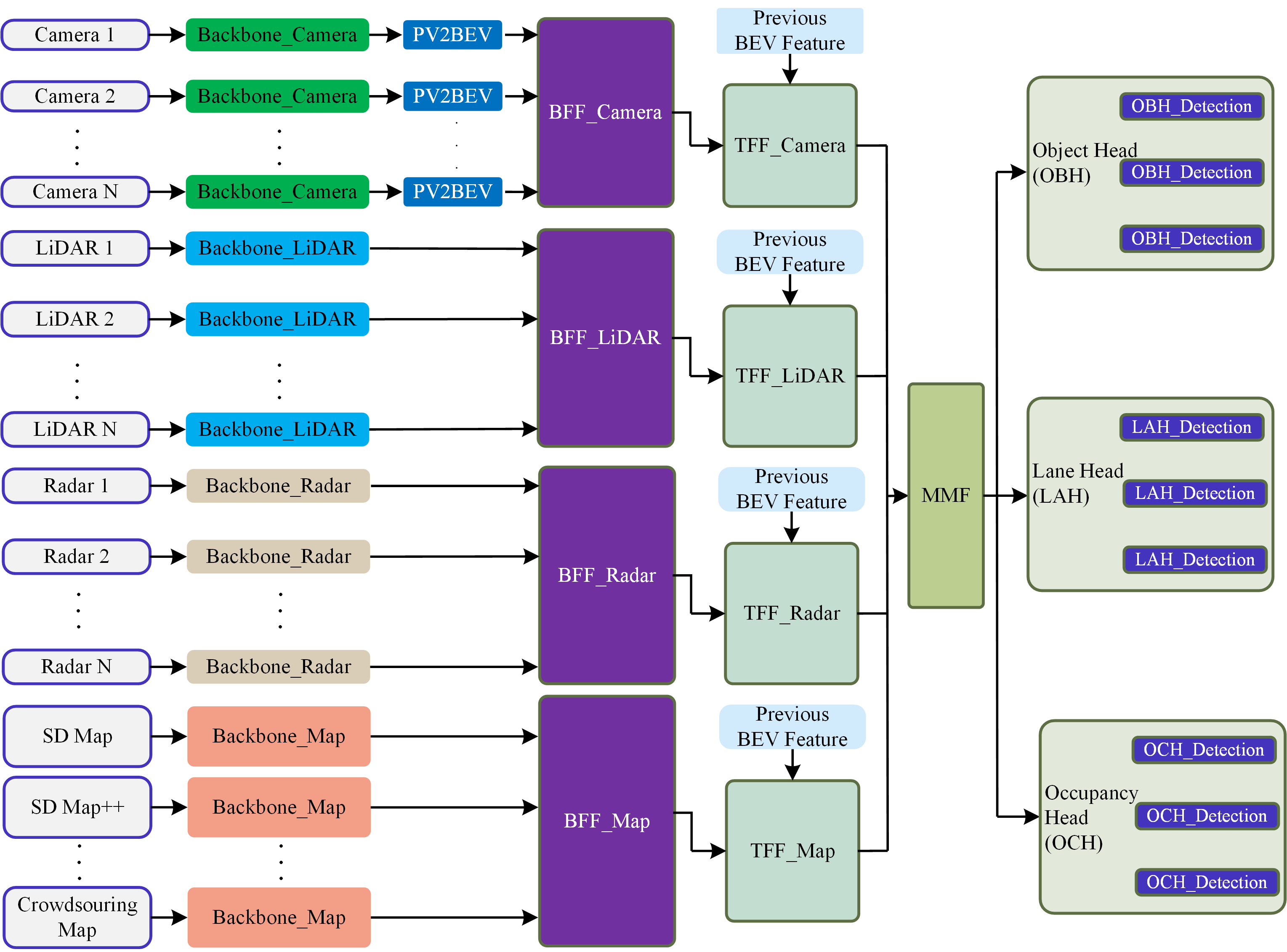}
  \caption{An overview of the proposed decoupled perception system for autonomous driving vehicles.}
  \label{fig:mps}
\end{figure*}

However, the balancing act between multiple concurrent tasks can also pose challenges. On one hand, it can lead to synergistic benefits, but on the other, it can cause a drop in accuracy due to the inherent trade-offs between different tasks \cite{wei2020combating}. Models underpinned by multi-task learning are tasked with extracting knowledge from various domains, a process that is significantly influenced by the shared parameters across tasks \cite{jeffares2024joint}. To fully leverage the benefits of multi-task learning, it becomes imperative to incorporate meaningful feature representations and implement balanced learning methodologies \cite{huang2022modality}. These strategies are crucial for striking the right balance, ensuring that each task receives the attention it requires for optimal performance while minimizing detrimental inter-task interference.

\section{Methodology}
In this section, we first introduce the details of the proposed hierarchical and decoupled perception scheme, which demonstrates the top-down design process and its advantages over traditional paradigm. Following this, we present the prototype design process of each component for the task of 3D object detection. Finally, we present our proposed MML method, which helps improve performance on individual model through jointly training and temporal updating of multiple common module weights.

\begin{table*}[!t]
\centering
\caption{Functional Module Library in Perception for Vehicle-Road-Cloud Integrated Architecture}
\begin{tabular}{c|c|c}
\noalign{\hrule height 1.5pt}
\textbf{Index} & \textbf{Module Name} & \textbf{Description} \\
\hline
1 & Backbone\_Camera & Image Feature Extraction \\
\hline
2 & PV2BEV & Image Feature Conversion from Perspective View to Bird's Eye View \\
\hline
3 & BFF\_Camera & Bird's Eye View Image Feature Fusion \\
\hline
4 & TFF\_Camera & Bird's Eye View Image Temporal Feature Fusion \\
\hline
5 & Backbone\_LiDAR & LiDAR Feature Extraction \\
\hline
6 & BFF\_LIDAR & LiDAR Feature Fusion \\
\hline
7 & TFF\_LIDAR & LiDAR Temporal Feature Fusion \\
\hline
8 & Backbone\_Radar & Millimeter Wave Radar Feature Extraction \\
\hline
9 & BFF\_Radar & Millimeter Wave Radar Feature Fusion \\
\hline
10 & TFF\_Radar & Millimeter Wave Radar Temporal Feature Fusion \\
\hline
11 & Backbone\_Map & Lightweight Map Feature Extraction \\
\hline
12 & BFF\_Map & Lightweight Map Feature Fusion \\
\hline
13 & TFF\_Map & Lightweight Map Temporal Feature Fusion \\
\hline
14 & MMF & Multi-Modal Feature Fusion \\
\hline
15 & OBH\_Detection & Obstacle Detection Head \\
\hline
16 & OBH\_Tracking & Obstacle Tracking Head \\
\hline
17 & OBH\_Prediction & Obstacle Prediction Head \\
\hline
18 & LAH\_Detection & Lane Detection Head \\
\hline
19 & LAH\_Tracking & Lane Tracking Head \\
\hline
20 & LAH\_Prediction & Lane Prediction Head \\
\hline
21 & OCH\_Detection & Occupancy Grid Detection Head \\
\hline
22 & OCH\_Tracking & Occupancy Grid Tracking Head \\
\hline
23 & OCH\_Prediction & Occupancy Grid Prediction Head \\
\hline
24 & FCT\_Camera\_Vehicle & Bird's Eye View Image Feature Compression and Transmission \\
\hline
25 & FCT\_LiDAR\_Vehicle & LiDAR Feature Compression and Transmission \\
\hline
26 & FCT\_Radar\_Vehicle & Millimeter Wave Radar Feature Compression and Transmission \\
\hline
27 & Backbone\_Camera\_Roadside & Roadside Image Feature Extraction \\
\hline
28 & PV2BEV\_Roadside & Roadside Image Feature Conversion from Perspective View to Bird's Eye View \\
\hline
29 & BFF\_Camera\_Roadside & Roadside Bird's Eye View Image Feature Fusion \\
\hline
30 & TFF\_Camera\_Roadside & Roadside Bird's Eye View Image Temporal Feature Fusion \\
\hline
31 & Backbone\_LiDAR\_Roadside & Roadside LiDAR Feature Extraction \\
\hline
32 & BFF\_LiDAR\_Roadside & Roadside LiDAR Feature Fusion \\
\hline
33 & TFF\_LiDAR\_Roadside & Roadside LiDAR Temporal Feature Fusion \\
\hline
34 & Backbone\_Radar\_Roadside & Roadside Millimeter Wave Radar Feature Extraction \\
\hline
35 & BFF\_Radar\_Roadside & Roadside Millimeter Wave Radar Feature Fusion \\
\hline
36 & TFF\_Radar\_Roadside & Roadside Millimeter Wave Radar Temporal Feature Fusion \\
\hline
37 & FCT\_Camera\_Roadside & Roadside Bird's Eye View Image Feature Compression and Transmission \\
\hline
38 & FCT\_LiDAR\_Roadside & Roadside LiDAR Feature Compression and Transmission \\
\hline
39 & FCT\_Radar\_Roadside & Roadside Millimeter Wave Radar Feature Compression and Transmission \\
\noalign{\hrule height 1.5pt}
\end{tabular}
\label{tab:mlibrary}
\end{table*}

\subsection{Hierarchical Perception Paradigm}
A high-level overview of our hierarchical and decoupled BEV perception learning scheme is depicted in Fig.~\ref{fig:hps}. The core innovation of hierarchical perception system is to provide a modular and easy-to-operate view of the BEV perception algorithm construction process, which is achieved by integrating pre-training and fine-tuning algorithms of functional module components, respectively.

In the context of an integrated vehicle-cloud-road architecture, a suite of 39 generic perception function modules is encapsulated, as delineated in Fig.~\ref{fig:mps} and Table \ref{tab:mlibrary}, representing a conceptual categorization. Building on single-vehicle intelligent perception, the vehicle-road-cloud integrated perception system primarily incorporates collaborative perception between vehicles and between vehicles and infrastructure. To build a library of numerical perception functional modules, a knowledge sharing training approach that integrates multi-module is employed. Initially, an exhaustive combination of perception function modules is pre-trained on the public vast dataset, thereby creating a repository of modular components. Subsequently, a user-friendly, graphical software interface is engineered to facilitate model construction through intuitive drag-and-drop operations. Moreover, the graphical software also provides a unified platform for training, inference, and fine-tuning.

To implement the design scheme outlined in Fig. \ref{fig:hps}, a phased and iterative approach is adopted. The phased implementation steps are organized based on the combination of key elements: sensor types and perception targets. This next subsection presents the design principles of a hierarchical and decoupled perception system for the task of 3D object detection, detailing its innovative aspects.

\begin{table*}[htbp]
\centering
\renewcommand{\arraystretch}{1.5}
\caption{Details of Different Model Configurations.}
\begin{tabular}{c|>{\centering\arraybackslash}p{2.5cm}|>{\centering\arraybackslash}p{2.5cm}|>{\centering\arraybackslash}p{3.8cm}|>{\centering\arraybackslash}p{2.5cm}|>{\centering\arraybackslash}p{1.5cm}|>{\centering\arraybackslash}p{1.5cm}}
\noalign{\hrule height 1.5pt}
Num. & Image Backbone & View Transform & Temporal Feature Fusion & Detection Head & Params(M) & FPS\\ 
\hline
1 &  ResNet-50 & SCA & TSA & 3D ODH & 33.57 & 16.9\\
2 &  ResNet-50 & SCA & RTF & 3D ODH & 33.97 & 17.5\\ 
3 &  ResNet-50 & GKT & TSA & 3D ODH & 33.47 & 16.9\\ 
4 &  ResNet-50 & GKT & RTF & 3D ODH & 33.87 & 17.0\\ 
5 &  VoVNetv2-99 & SCA & TSA & 3D ODH & 79.33 & 11.9\\ 
6 &  VoVNetv2-99 & SCA & RTF & 3D ODH & 79.72 & 12.1\\ 
7 &  VoVNetv2-99 & GKT & TSA & 3D ODH & 79.23 & 11.9\\ 
8 &  VoVNetv2-99 & GKT & RTF & 3D ODH & 79.62 & 12.1\\ 
\noalign{\hrule height 1.5pt}
\end{tabular}
\label{tab:allmodels}
\end{table*}

\subsection{Prototype Design for Vision-centric 3D Object Detection}
3D object detection is a cornerstone of the autonomous driving perception system, providing the vehicle with the ability to localize objects in three-dimensional space and estimate their dimensions and orientation \cite{qian20223d}. This capability is crucial for tasks such as collision avoidance and motion planning. The field has seen a proliferation of methods that leverage different types of sensors and learning paradigms \cite{lin2023sparse4d}. At this stage, we select the task of pure vision-based 3D object detection to realize a prototype system. This section provides a detailed description of the modular 3D object detection algorithm construction.



Following the hierarchical and decoupled modular designing thoughts, the proposed  3D object detection method integrates a suite of functional components:  (1) An image-view feature extractor is employed to capture and encode visual features effectively. (2) Subsequently, a view transformer is utilized to translate these encoded features from a perspective view to the BEV perspective. (3) A temporal feature fusion module further enhances the feature representation by integrating temporal information. (4) Finally, a task-specific detection head is utilized to perform 3D object detection.

\subsubsection{\textbf{Feature Extraction}}
Specifically, the main function of image feature extraction module is to extract hierarchical and abstract visual feature from input images to form a comprehensive visual representation that capture essential visual information such as edges, textures, shapes, and semantic content. This process can be expressed as the following: at timestamp $t$, multi-camera images are fed into a backbone network, yielding features $\mathbf{F}^t = \{\mathbf{F}^t_i\}_{N_{\text{view}} i=1}$ for different camera views, with $\mathbf{F}^t_i$ being the features from the $i$-th view and $N_{\text{view}}$ being the total number of views.

To addressing the diverse computational capabilities of platforms, two types of backbone networks with varying parameter counts and architectural complexities have been strategically chosen, including ResNet series model \cite{he2016resnet} and VoVNet series model \cite{lee2020centermask,lee2019energy}. Renowned for their performance, these networks provide users with the flexibility to choose the optimal network configuration that aligns with specific application contexts and computational needs. 

In the construction of the prototype system, we employ two representative backbone networks. The first one is ResNet-50, with 25.6 million parameters and a computational load of roughly 3.8 GFLOPs. ResNet-50 strikes an balance between efficiency and capability, fitting well with platforms of moderate computational power. The other one is VoVNetv2-99. VoVNetv2-99 stands out for its dense connectivity, engineered to enhance memory use and speed up inference while maintaining high performance. Boasting a significant parameter count of 96.9 million, it's recognized as an ideal backbone for scenarios where computational resources are plentiful and high-performance demands are prevalent.

\subsubsection{\textbf{Image View Transformation}}
The image view transformation module we applied contains two approaches: Spatial Cross Attention (SCA) \cite{bevformer} and Geometry-guided Kernel Transformer (GKT) \cite{chen2022efficient}, both fundamentally grounded in transformer-based approaches to model the perspective transformation in a data-driven fashion. 

The SCA mechanism is designed to capture the inter-relation of features across different perspectives. Initially, perspective features and the preliminarily transformed BEV features are linearly mapped onto a unified feature space. A set of grid-shaped learnable parameters $\mathbf{Q} \in \mathbb{R}^{H \times W \times C}$ are predefined as the queries for BEV representation, termed BEV Queries, where $H, W$ represent the spatial dimensions of the BEV plane, and $C$ is the channel dimension. Each grid cell in the BEV plane maps to a real-world size of $s$ meters. Then, each BEV query $\mathbf{Q}_p$ is lifted to a pillar-like query, samples $N_{\text{ref}}$ 3D reference points from the pillar, and projects these onto 2D views. A projection function $P(p, i, j)$ is defined to calculate the 2D points on the $i$-th view image from the $j$-th 3D reference point corresponding to the BEV query $\mathbf{Q}_p$ located at $p = (x, y)$:
\begin{align}
\begin{split}
P(p, i, j) = (x_{ij}, y_{ij})
\end{split}
\end{align}
The real-world location $(x', y')$ corresponding to the BEV query $\mathbf{Q}_p$ is calculated as:
\begin{align}
\begin{split}
x' = (x - \frac{W}{2}) \times s; \quad y' = (y - \frac{H}{2}) \times s 
\end{split}
\end{align}
A set of anchor heights $\{z'_j\}_{j=1}^{N_{\text{ref}}}$ is predefined to capture cues at different heights, and the 3D reference points $(x', y', z'_j)$ are projected onto different camera views using the camera's projection matrix $\mathbf{T}_i$.

Attention weights are then computed to indicate the correlation between features at various locations. These weights are subsequently utilized for a weighted summation of features, achieving feature fusion. The SCA output is a weighted sum of the sampled features from the regions of interest in the camera views that are hit by the projected points:
\begin{align}
\begin{split}
\text{SCA}(\mathbf{Q}_p, \mathbf{F}^t) = \frac{1}{|V_{\text{hit}}|} \sum_{i \in V_{\text{hit}}} \sum_{j=1}^{N_{\text{ref}}} \text{DeformAttn}(\mathbf{Q}_p, P(p, i, j), \mathbf{F}^t_i)
\end{split}
\end{align}
where $V_{\text{hit}}$ is a set of camera views that contain the projected 2D points (hit views). For each BEV query, the deformable attention mechanism computes the interaction with the 2D view features, generating attention weights that are used to weight and sum the sampled features:
\begin{align}
\begin{split}
\text{DeformAttn}(\mathbf{Q}_p, \mathbf{p}, \mathbf{F}^t_i) = \sum_{k=1}^{N_{\text{key}}} A_{p,k} \mathbf{W}' \mathbf{F}^t_i(\mathbf{p} + \Delta \mathbf{p}_{p,k})
\end{split}
\end{align}
where $A_{p,k}$ represents the predicted attention weight, normalized such that $\sum_{k=1}^{N_{\text{key}}} A_{p,k} = 1$, $\mathbf{W}'$ is the learnable weight matrix, and $\Delta \mathbf{p}_{p,k}$ are the predicted offsets.

On the other hand, the GKT presents a novel image view transformation mechanism, which optimizes the feature transformation process by amalgamating geometric transformations with transformer. GKT utilizes geometric priors to guide the transformers, focusing on discriminative regions for generating BEV representation with surrounding-view image features. Geometric information, such as depth maps and camera parameters, is initially harnessed to generate geometry-guided convolutional kernels. With the camera's extrinsic and intrinsic parameters, each BEV grid is projected to approximate 2D positions in the multi-view and multi-scale feature maps.
\begin{align}
\begin{split}
Q_{sv}^i = K_{sv} \cdot R_{tsv} \cdot T_{tsv} \cdot P_{sv}^i \\
\end{split}
\end{align}
\noindent where \( Q_{sv}^i \) represents the 2D positions in various views and scales, \( K_{sv} \) is the camera projection matrix, \( R_{tsv} \) and \( T_{tsv} \) are the rotation and translation from the BEV to the camera coordinate system, and \( P_{sv}^i \) is the 3D coordinate of the BEV grid. These coordinates are then rounded to the nearest integer coordinates which are used to localize the region of interest on the 2D feature map. Next, the nearby $K_h \times K_w$ sized kernel region is unfolded around these a priori locations by GKT, facilitating the interaction between BEV queries and the corresponding unfolded features to generate BEV representations. It is based on kernel-wise attention, making it significantly efficient, and employs LUT indexing for fast inference. Importantly, GKT demonstrates robustness to camera deviations, contributing to a more stable and reliable view transformation.
\begin{figure}[!t]
\centering
\includegraphics[width=\linewidth]{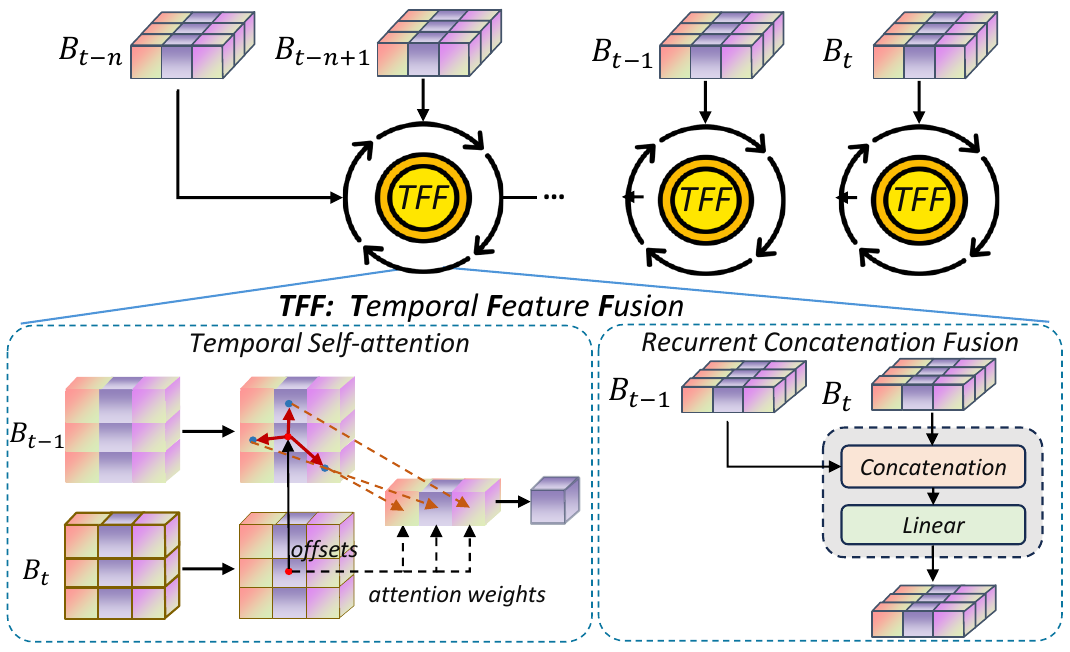}
\caption{Architecture of temporal feature fusion modules.}
\label{fig:tff}
\end{figure}
\subsubsection{\textbf{Temporal Feature Fusion}}
Temporal cues from historical frames provide additional information for 3D perception in autonomous driving. The Temporal Feature Fusion (TFF) module is used to integrate temporal features from BEV imagery, enhancing the understanding of dynamic environments. The process of temporal feature fusion involves three critical steps. First, a selection process of preceding frames determines the temporal extent of the fusion. Specifically, we sample the current frame with three history key frames and the initial frame is fused with a copy of itself due to the absence of a prior frame for comparison. Second, spatio-temporal alignment is executed by adjusting the features of the preceding frames in accordance with the ego-motion, ensuring they align with the current frame's features within a unified coordinate system. This alignment is crucial for precise feature integration. The final step is the integration of temporal information. As depicted in Fig \ref{fig:tff}, the Temporal Self-Attention (TSA) module and the Recurrent Concatenation Fusion (RCF) module are integrated in our hierarchical 3D object detector.

\begin{figure*}[!t]
\centering
\includegraphics[width=\textwidth]{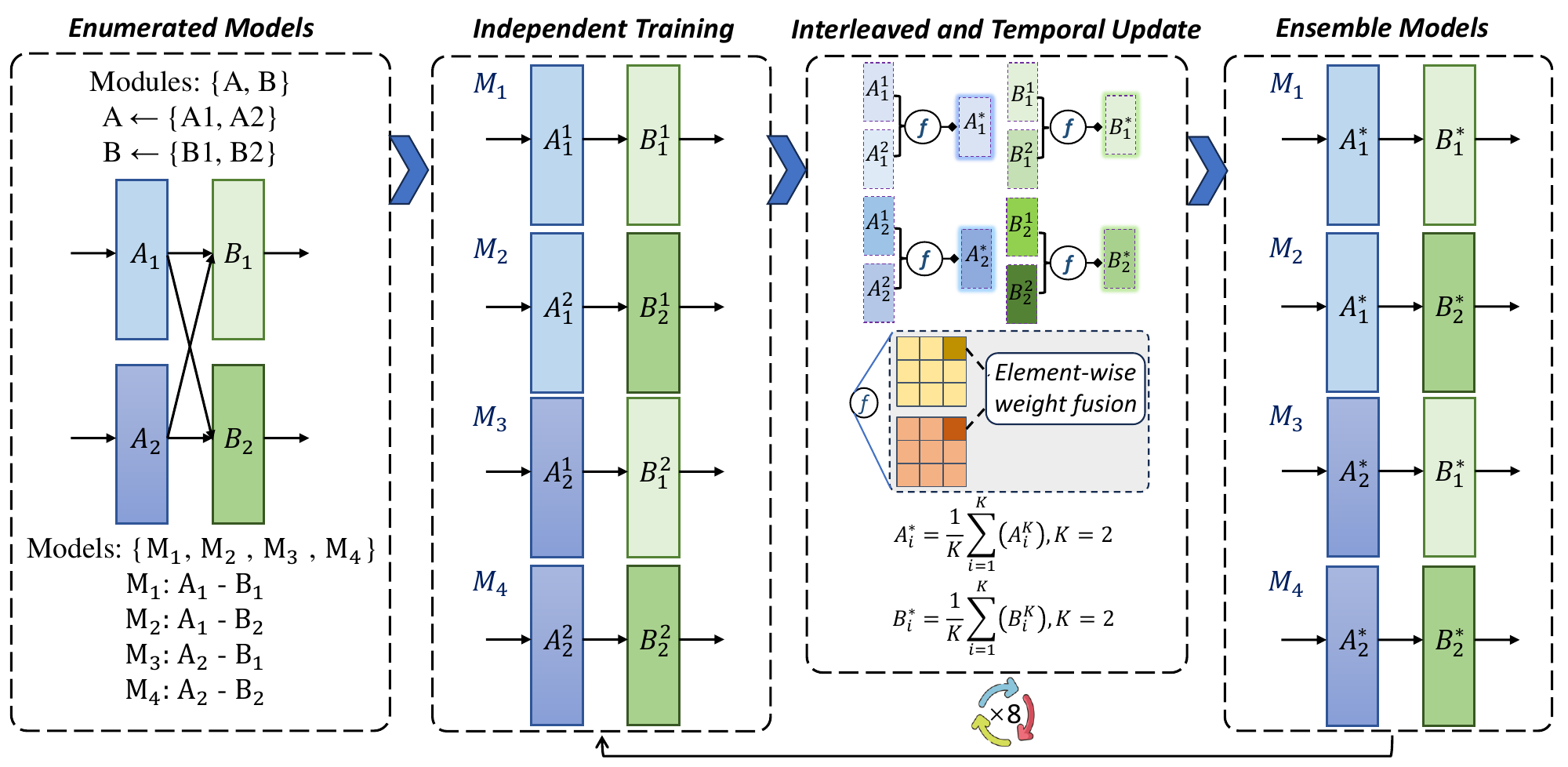}
\caption{Sketch of the proposed Multi-Module Learning pipeline. Taking a 2x2 combination as an example to illustrate the proposed pre-training process for functional modules. First, we conduct a single mini-epoch training session with a mini-epoch size of 3 for each of the various combination models. After this round of training, we perform a parameter fusion and update for the weights of the modules that are common across all models. Then, we continue with further training iterations for optimization. This process is halted after reaching the preset maximum number of training epochs (set to 8 in the experiment), at which point we obtain the final weights for the functional modules.}
\label{fig:MML-sketch}
\end{figure*}


The core of the TSA is the self-attention mechanism, which computes the attention weights based on the interactions between the BEV queries and the aligned historical BEV features. The TSA module integrates the spatially and temporally aligned preceding and current BEV features by applying Deformable Attention to each, followed by an arithmetic mean fusion on the $H \times W$ plane. Given a set of BEV queries \( \mathbf{Q} \) and the corresponding historical BEV features \( \mathbf{B}_{t-1} \), the TSA computes a self-attention mechanism that captures temporal dependencies as shown in Eq.\eqref{eq:tsa}. 
\begin{align}
\begin{split}
\text{TSA}(\mathbf{Q}_p, \{\mathbf{Q}, \mathbf{B}_{t-1}'\}) = \sum_{\mathbf{V} \in \{\mathbf{Q}, \mathbf{B}_{t-1}'\}} \text{DeformAttn}(\mathbf{Q}_p, \mathbf{p}, \mathbf{V})
\end{split}
\label{eq:tsa}
\end{align}

The TSA module operates on the premise of utilizing historical BEV features to enhance the current representation. While the RCF module is designed based on a combination of cascaded operations and linear layers, fuses the aligned preceding and current BEV features in a manner that synthesizes information across temporal dimensions.

\subsubsection{\textbf{Detection Head}}
We take BEVFormer-tiny as our baseline and adopt the modified 3D detection head based on Deformable DETR~\cite{zhu2020deformable} to obtain final prediction results. The detection head consists of 6 decoder layers, i.g. interleaved self-attention and cross-attention layers. By utilizing single-scale BEV features $B_t$ as input, the decoder is capable of predicting 3D bounding boxes and velocity in an end-to-end manner. The final loss is composed of two components: a classification $\mathcal{L}_{cls}$ and a regression loss $\mathcal{L}_{reg}$, which is defined as below: 
\begin{equation}
\begin{split} 
\mathbf{Loss} = \mathcal{L}_{cls} + \mathcal{L}_{reg}
\end{split}
\end{equation}
where $\mathcal{L}_{cls}$ is focal loss~\cite{lin2017focal} for target classification and $\mathcal{L}_{reg}$ is $\mathcal{L}_{1}$ loss for target localization.

\subsection{Multi-Module Learning Framework}
\label{ssec:MML}
Once the perception functional component are constructed, we introduce a novel \textbf{M}ulti-\textbf{M}odule \textbf{L}earning (MML) framework in the pre-training stage. The pre-training regimen involves an exhaustive training of each potential module combination, ensuring that eachperceptual model is endowed with compatible weights for both upstream and downstream functional counterparts. To delineate MML comprehensively, we initially provide a definition of MML.

\begin{algorithm}[!t]
    \SetKwInput{Input}{Input}
    \SetKwInput{Output}{Output}
    \SetKwInput{Parameter}{Parameter}
    \SetKwInput{Initialize}{Initialize}
    \Input{Trained model weights : $\mathcal{\theta} = \{ \mathbf{\theta}_1, \mathbf{\theta}_2, \ldots, \mathbf{\theta}_N \}$ \\
     Number of different models: $N$ \\
     Functional modules: $m=\{m_{EF}, m_{PV2BEV}, m_{TFF}, m_{HEAD}\}$ \\
     $\mathbf{\theta}_\text{FE} \gets \{\},$ 
     $\mathbf{\theta}_\text{PV2BEV} \gets \{\},$ 
     $\mathbf{\theta}_\text{TFF} \gets \{\}$
     $\mathbf{\theta}_\text{HEAD} \gets \{\}$
     }
    \Initialize{Load trained parameters in mini epochs: $\theta$}
    \For{$i = 1$ to $N$}{
        \If{$\text{m}$ in $\theta_{i}$}{
        \If{$m_{EF}$ in $\text{m}$}{
            $\mathbf{\theta}_\text{FE} \gets \mathbf{\theta}_\text{FE} \cup \{\theta_i\}$
        }
        \If{$m_{PV2BEV}$ in $\text{m}$}{
            $\mathbf{\theta}_\text{PV2BEV} \gets \mathbf{\theta}_\text{PV2BEV} \cup \{\theta_i\}$
        }
        \If{$m_{TFF}$ in $\text{m}$}{
            $\mathbf{\theta}_\text{TFF} \gets \mathbf{\theta}_\text{TFF} \cup \{\theta_i\}$
        }
        \If{$m_{HEAD}$ in $\text{m}$}{
            $\mathbf{\theta}_\text{HEAD} \gets \mathbf{\theta}_\text{HEAD} \cup \{\theta_i\}$
        }
       Update model parameters: $\mathbf{\theta}_\text{FE} \leftarrow average(\mathbf{\theta}_\text{FE})$
       $\mathbf{\theta}_\text{PV2BEV} \leftarrow average(\mathbf{\theta}_\text{PV2BEV})$
      $\mathbf{\theta}_\text{TFF} \leftarrow average(\mathbf{\theta}_\text{TFF})$
      $\mathbf{\theta}_\text{HEAD} \leftarrow average(\mathbf{\theta}_\text{HEAD})$
      }     
    }
    \Output{Assemble Weights: $\mathbf{\theta}_\text{FE}, \mathbf{\theta}_\text{PV2BEV}, \mathbf{\theta}_\text{TFF},\mathbf{\theta}_\text{HEAD}$}
    \caption{Averaging Module Assemble Strategy}
\label{alg:mml} 
\end{algorithm}

\textbf{Definition}: Given {$m$} functional modules {$M_i$}, where all the modules or a subset of them are related, multi-module learning aims to jointly optimize these {$m$} modules to enhance the performance of a model for the common task by leveraging the knowledge contained in all model ensembles.


As shown in Fig.\ref{fig:MML-sketch}, a simple $2\times2$ combination is taken as an illustrative example to detail the proposed MML pipeline. At first, we enumerate all combinations of these modules. Two different A-functional modules $\{A_1,A_2\}$ and two different B-functional modules $\{B_1,B_2\}$ form a total of 4 combination models $\{M_1,M_2,M_3,M_4\}$. The training workflow is delineated as follows. \textbf{Step 1: Independent Training.} Each model $\{M_i\}$ is trained using a traditional end-to-end training paradigm for mini-epoch=3, to obtain the initial weights for the first round. \textbf{Step 2: Interleaved and Temporal Update.} The main computational process involves loading the model weight, traversing the model weight dictionary, and match the common modules according to the dictionary keys. Subsequently, the weights of the same module from different models are averaged and updated as Eq.\eqref{eq:mave}, leveraging a averaging module assemble strategy. \textbf{Step 3: Ensemble models.} Once the average value has been computed, the weights of each module are reloaded to their respective models, resulting in an update to the model parameters across the board. The aforementioned three steps are collectively defined as one round in this iterative process. The propsed MML framework iterates cyclically for eight rounds, ultimately yielding the final functional module weights.
\begin{equation}
\begin{split} 
A_i^* = \frac{1}{K} \sum_{i=1}^{K} \left( A_i^K \right), \quad K = 2 \\
B_i^* = \frac{1}{K} \sum_{i=1}^{K} \left( B_i^K \right), \quad K = 2 \\
\end{split}
\label{eq:mave}
\end{equation}

In the context of vision centric 3D object detection, the entire BEV perception model is divided into four parts and the calculation of the total number of combinations $N_{total}$ for the model is as $ N_{total} = {N_{FE}}\times{N_{PV2BEV}}\times{N_{TFF}}\times{N_{HEAD}}$. $N_{*}$ represents the number of modules contained in a certain part and the relevant functional module can be represented as $m(x,\theta)$, $\theta \in \mathbb{R}^{d}$, where $x$ denotes the input data and $\theta$ represents the weight parameters. Thus, the proposed modular assembly method can be formulated as Eq.\eqref{eq:hdm}.
\begin{align}
\begin{split}
HDP(x,\theta) &= \left\{ m_{FE}(x,\theta_{F}),  m_{PV2BEV}(f_{1},\theta_{P}), \right. \\
  & \left. m_{TFF}(f_{2},\theta_{T}),,  m_{HEAD}(f_{3},\theta_{H})\right \}
\end{split}
\label{eq:hdm}
\end{align}
where $m_{*}$ represents different functional modules, and $f_{*}$ represents the features input to different modules. As shown in Algorithm \ref{alg:mml}, during the model weight fusion and updating phase, we iterate over all $N$ weights trained for a mini-epoch. If the weight $\theta$ contains the corresponding functional module weight $\theta_{m}$, it will be saved and averaged with all weights that containing the same functional module as Eq.\eqref{eq:ave}.
\begin{equation}
\begin{aligned}
\theta_{FE} = \frac{1}{N_{F}}  {\textstyle \sum_{i\in N_{F} }^{}} \theta _{i} \\
\theta_{PV2BEV} = \frac{1}{N_{P}}  {\textstyle \sum_{i\in N_{P} }^{}} \theta _{j} \\ 
\theta_{TFF} = \frac{1}{N_{T}}  {\textstyle \sum_{i\in N_{T} }^{}} \theta _{k} \\
\theta_{HEAD} = \frac{1}{N_{H}}  {\textstyle \sum_{i\in N_{H} }^{}} \theta _{m} \\
\end{aligned}
\label{eq:ave}
\end{equation}
$N_{*}$ represents the number of models containing a certain module, which is computed by Eq.\eqref{eq:num}.
\begin{equation}
\begin{aligned}
N_{F}=N_{PV2BEV}\times N_{TFF}\times N_{HEAD} \\
N_{P}=N_{FE}\times N_{TFF}\times N_{HEAD} \\ 
N_{T}=N_{FE}\times N_{PV2BEV}\times N_{HEAD} \\
N_{H}=N_{FE}\times N_{PV2BEV}\times N_{TFF} \\
\end{aligned}
\label{eq:num}
\end{equation}


\textit{Difference with Existing MTL paradigm.} Firstly, the proposed MML is directed towards the development of a library of perception functional modules. Utilizing MML, a variety of related model architectures are trained concurrently, achieving collective performance gains. This approach is pertinent not only to single-task learning but is also adaptable to multi-task learning contexts. Secondly, MML employs a strategy of soft parameter sharing to achieve knowledge transfer and provides a flexible and loosely coupled architecture for joint learning of multiple models.  Our method maintain the unique structure of individual model, varying from the direct sharing of substantial network layers in mainstream MTL framework. Thirdly, the proposd MML acknowledges that distinct models necessitate the capacity of intrinsic feature space modeling.  By bestowing each model with a dedicated feature representational learning space, the overall performance is augmented. 


\begin{table*}[!t]
\centering
\renewcommand{\arraystretch}{1.3}
\caption{MML-Ave achieves consistent improvements for different ensemble models. With 24 epochs pre-training on nuScenes train dataset and 24 epochs fine-tuning on 10\% nuScenes validation dataset for different ensemble model.}
\begin{tabular}{l  c c |c c| c c c c c c  }
\toprule 
\toprule
Method &  Method & Epoch & NDS$\uparrow$  & mAP$\uparrow$  & mATE$\downarrow$     & mASE$\downarrow$     & mAOE$\downarrow$     & mAVE$\downarrow$    & mAAE$\downarrow$    \\
\midrule
\multirow{2}{*}{Res50-SCA-TSA} & Baseline & 24 & 0.2141 & 0.3662 & 0.8521 & 0.2598 & 0.4868 & 0.5774 & 0.2319 \\
 & MML-Ave & 24 & \textbf{0.2229} & \textbf{0.3798} & \textbf{0.8240} & \textbf{0.2491} & \textbf{0.4655} & \textbf{0.5659} & \textbf{0.2122} \\
\midrule
\multirow{2}{*}{Res50-SCA-RCF} & Baseline & 24 & 0.2103 & 0.3553 & 0.8360 & 0.2564 & 0.5022 & 0.6650 & 0.2389 \\
 & MML-Ave & 24 & \textbf{0.2158} & \textbf{0.3704} & \textbf{0.8307} & \textbf{0.2549} & \textbf{0.4385} & \textbf{0.6275} & \textbf{0.2230} \\
\midrule
\multirow{2}{*}{Res50-GKT-TSA} & Baseline & 24 & 0.1842 & 0.3230 & 0.8807 & 0.2741 & 0.5142 & 0.7534 & 0.2685 \\
 & MML-Ave & 24 & \textbf{0.2130} & \textbf{0.3697} & \textbf{0.8310} & \textbf{0.2565} & \textbf{0.4496} & \textbf{0.6069} & \textbf{0.2238} \\
\midrule
\multirow{2}{*}{Res50-GKT-RCF} & Baseline & 24 & 0.1970 & 0.3490 & 0.8636 & 0.2618 & 0.4759 & 0.6635 & 0.2303 \\
 & MML-Ave & 24 & \textbf{0.2087} & \textbf{0.3637} & \textbf{0.8387} & \textbf{0.2523} & \textbf{0.4430} & \textbf{0.6499} & \textbf{0.2224} \\
\midrule
\multirow{2}{*}{VoV-SCA-TSA} & Baseline & 24 & 0.3301 & 0.4661 & 0.7265 & 0.2503 & 0.3168 & \textbf{0.4955} & \textbf{0.2004} \\
& MML-Ave & 24 & \textbf{0.3336} & \textbf{0.4699} & \textbf{0.7231} & \textbf{0.2466} & \textbf{0.2976} & 0.4983 & 0.2034 \\
\midrule
\multirow{2}{*}{VoV-SCA-RCF} & Baseline & 24 & 0.3198 & 0.4449 & \textbf{0.7260} & 0.2496 & 0.3328 & 0.6256 & 0.2153 \\
 & MML-Ave & 24 & \textbf{0.3222} & \textbf{0.4531} & 0.7356 & \textbf{0.2446} & \textbf{0.3103} & \textbf{0.5833} & \textbf{0.2057} \\
\midrule
\multirow{2}{*}{VoV-GKT-TSA} & Baseline & 24 & 0.3094 & 0.4437 & 0.7455 & 0.2521 & 0.3332 & 0.5680 & 0.2110 \\
& MML-Ave & 24 &\textbf{0.3252} & \textbf{0.4606} & \textbf{0.7381} & \textbf{0.2437} & \textbf{0.3062} & \textbf{0.5256} & \textbf{0.2061} \\
\midrule
\multirow{2}{*}{VoV-GKT-RCF} & Baseline & 24 & 0.3006 & 0.4299 & 0.7579 & 0.2510 & 0.3343 & 0.6530 & 0.2078 \\
 & MML-Ave & 24 & \textbf{0.3107} & \textbf{0.4451} & \textbf{0.7331} & \textbf{0.2447} & \textbf{0.3131} & \textbf{0.6103} & \textbf{0.2014} \\
\bottomrule
\bottomrule
\end{tabular}
\label{tab:ft_allmetrics}
\end{table*}

\section{EXPERIMENT}

\subsection{Implementation Details}
\textbf{Dataset and Metrics.} We conduct experiments on nuScenes \cite{nuscenes}, one of the most challenging public autonomous driving datasets. It is composed of 1000 multi-modal videos, each extending approximately 20 seconds with keyframes captured at 2 Hz intervals. The nuScenes dataset provides a wealth of sensor data, which ensures a complete 360-degree field of view through images from six cameras per sample. Each sample consists of RGB images from 6 cameras in the front, front-left, front-right, back-left, back-right and back directions. The nuScenes dataset is meticulously partitioned into 700 videos for training, 150 for validation, and 150 for testing purposes. For the detection task, there are 18538 annotated 3D bounding boxes from 10 categories. Additionally, we use the official evaluation metrics for evaluation. The mean average precision (mAP) of nuScenes is computed using the center distance on the ground plane rather than the 3D Intersection over Union (IoU) to match the predicted results and ground truth. The nuScenes metrics also contain 5 types of true positive metrics (TP metrics), including ATE, ASE, AOE, AVE, and AAE for measuring translation, scale, orientation, velocity, and attribute errors, respectively. The official nuScenes devkit also defines a nuScenes detection score (NDS)  to capture all aspects of the nuScenes detection tasks.

\textbf{Training Details.}
The experiments were conducted by implementing Python 3.8 and Pytorch 1.9.1. The corresponding codes are executed on four NVIDIA Tesla A100 GPUs with 40GB memory and with a batch size of 4. The initial learning rate was set to $2\times10^{-4}$ for training and dynamically updated the learning rate using the cosine strategy during the training process. All the hyper-parameters and settings are set as default following previous work \cite{bevformer} except the model parameters. Drawing inspiration from prior work in the field, the ResNet50 series models are initialized with weights from the FCOS3D~\cite{wang2021fcos3d} checkpoint and the VoVNetv2-99 series models are initialized with weights from the DD3D~\cite{park2021dd3d} checkpoint. In the fine-tuning phase, all the pre-trained weights are used to initialize the object detection model. 


\begin{figure*}[]
    \centering
    \includegraphics[width=\textwidth]{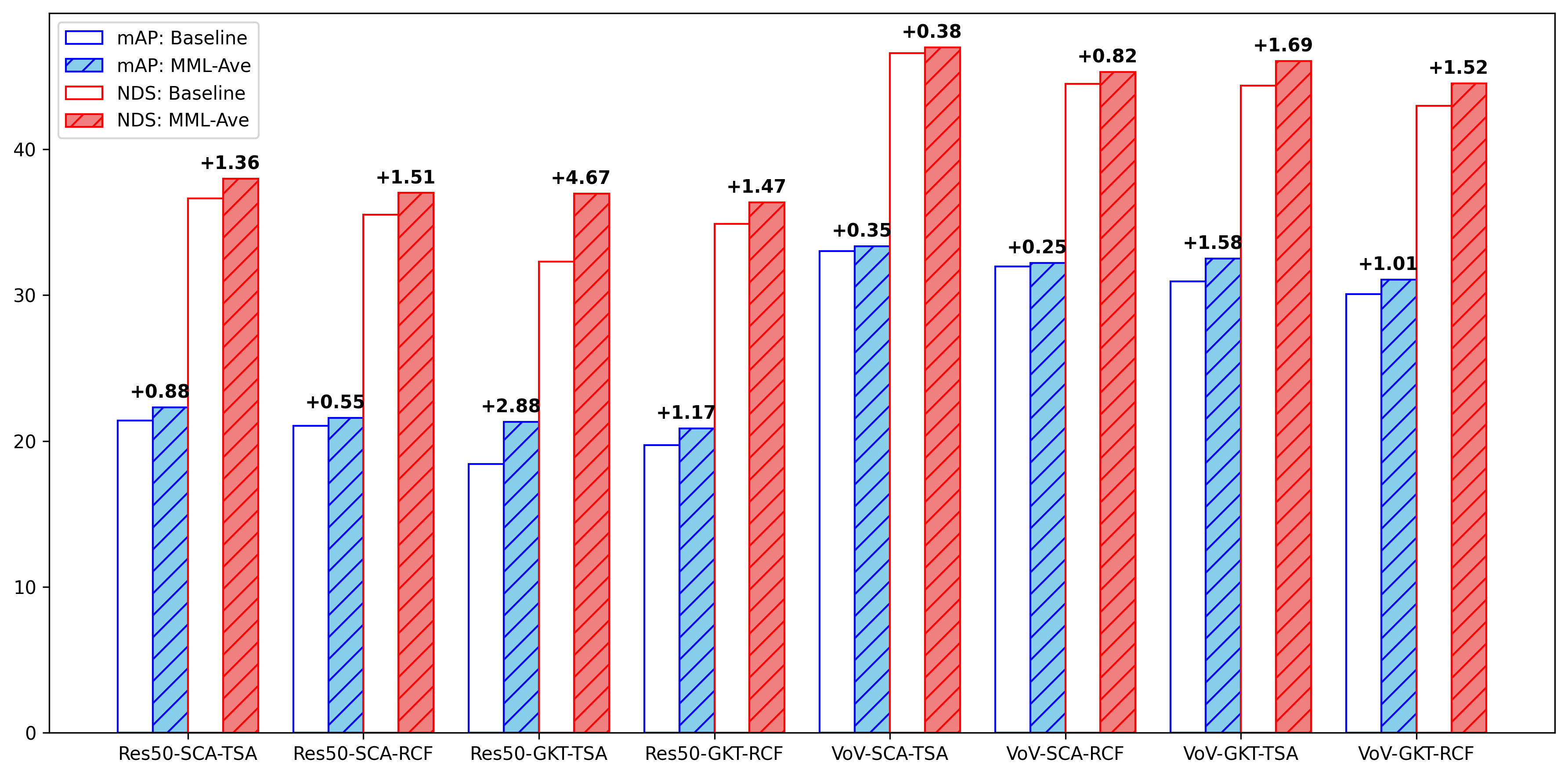}
    \caption{Comparison results between the traditional training approach and the proposed MML method on 90\% of the nuScenes validation dataset for different ensemble models.}
\label{fig:ftres10}
\end{figure*}

\begin{figure*}[htbp!]
    \centering
    \includegraphics[width=\textwidth]{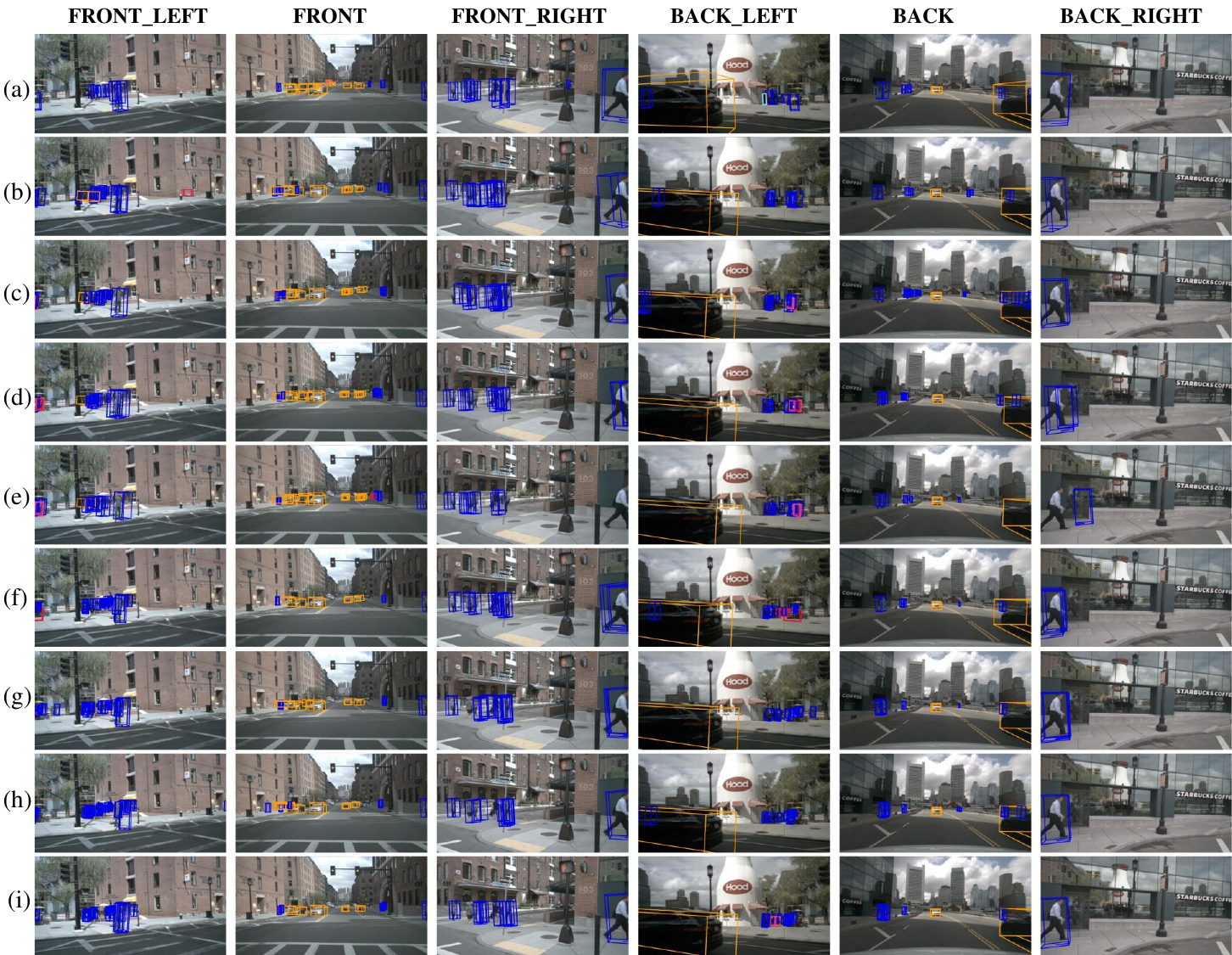}
    \caption{Visual comparison results on nuScenes val set. We show the 3D bboxes predictions in multi-camera images. (a): Ground Truth; (b)$\sim$(i): Prediction results of all the 8 models enumerated in Table~\ref{tab:allmodels}.}
    \label{fig:vis}
\end{figure*}

\begin{figure*}[htbp!]
    \centering
    \includegraphics[width=0.88\textwidth]{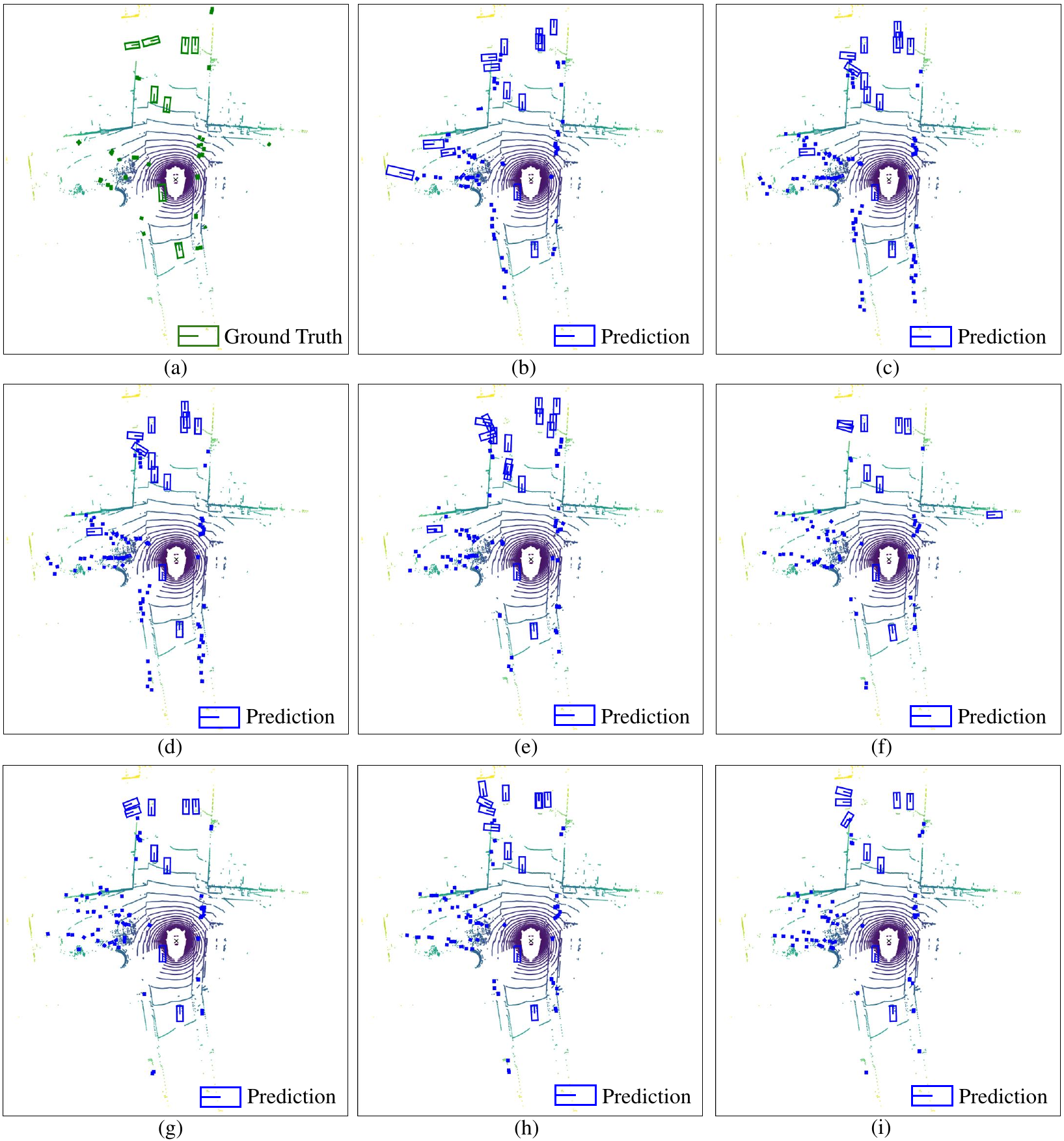}
    \caption{Visual comparison results on nuScenes val set. We show the 3D bboxes predictions in BEV and overlay the point clouds from lidar top. (a): Ground Truth; (b)$\sim$(i): Prediction results of all the 8 models enumerated in Table~\ref{tab:allmodels}.}
    \label{fig:vis_lidar}
\end{figure*}

\subsection{Comparative Results}

To demonstrate the effectiveness and generalization of our approach, we leverage the proposed MML with the averaging module assemble strategy, termed MML-Ave, to pretrain all the 8 composite pattern models, as demonstrated in Table \ref{tab:allmodels}. We random 10\% samples from nuScenes validation dataset as the fine-tuning data and the remaining 90\% data are used as test data. Table {\ref{tab:ft_allmetrics} llustrates the performance comparison achieved by our MML-Ave method when applied to various ensemble models in the context of the 3D object detection. All the models underwent a pre-training phase consisting of 24 epochs on the nuScenes training dataset, followed by a fine-tuning phase of another 24 epochs on 10\% nuScenes validation dataset. The results indicate that across all tested ensemble models, the application of MML-Ave leads to consistent improvements in both mAP and NDS. Moreover, Fig.\ref{fig:ftres10} further visualizes the comparative experimental results through data histograms. For each model, Fig.\ref{fig:ftres10} provides the baseline performance, the performance after applying the MML-Ave method, and the absolute improvement. The corresponding values of $\Delta$mAP and $\Delta$NDS are placed above each group of data as the relevant annotations in Fig.\ref{fig:ftres10}. 



As depicted in Fig.\ref{fig:ftres10}, the Res50-GKT-TSA model shows the most significant improvement, with a $\Delta$mAP of 2.9\% and a $\Delta$NDS of 4.7\%. Similarly, VoVNet based models also demonstrate notable improvements, particularly the VoV-GKT-TSA model with a $\Delta$mAP of 1.6\% and a $\Delta$NDS of 1.7\%. While VoVNet models generally outperform ResNet-50 models in absolute terms, the relative improvements are more substantial in the ResNet-50 based models. This indicates that these models benefit more from the integration of MML-Ave. In summary, all the experimental results presented above demonstrate the efficacy of the MML-Ave method in optimizing the performance of different ensemble models on the nuScenes dataset, as evidenced by the improvements in both mAP and NDS metrics. The results underscore the potential of MML-Ave approach to significantly enhance the generalization capability in 3D object detection tasks. 

As depicted in Fig.\ref{fig:vis}, the first row shows the ground truth, and the second to eighth rows are the predictions of all the eight models. The first to sixth columns are visualizations of the 6 views. Moreover, we also present visualization results of the bounding boxes in the BEV and overlay the point clouds from lidar top in Fig.\ref{fig:vis_lidar}.

\begin{table*}[htbp]
\centering
\renewcommand{\arraystretch}{1.5}
\caption{Ablation study of the weight essemble method, we report performance on 90\% Nuscenes Validation Datset.}
 \begin{tabular}{c >{\centering\arraybackslash}p{1.0cm}>{\centering\arraybackslash}p{1.5cm}>{\centering\arraybackslash}p{1.5cm}>{\centering\arraybackslash}p{1.5cm}>{\centering\arraybackslash}p{1.0cm}>{\centering\arraybackslash}p{1.5cm}>{\centering\arraybackslash}p{1.5cm}>{\centering\arraybackslash}p{1.5cm}}
\toprule
\toprule
\multirow{2}{*}{Model} & \multicolumn{4}{c}{mAP  $\uparrow$ (\%)} & \multicolumn{4}{c}{NDS $\uparrow$ (\%)} \\
\cline{2-9}
 & Baseline & MML-Average & MML-Softmax & MML-Greedy & Baseline & MML-Average & MML-Softmax & MML-Greedy \\
\hline
Res50-SCA-TSA & 21.41 & \textbf{22.29} & 20.31 & 18.83 & 36.62 & \textbf{37.98} & 33.92 & 32.80 \\
Res50-SCA-RTF & 21.03 & \textbf{21.58} & 19.34 & 18.53 & 35.53 & \textbf{37.04} & 32.39 & 32.06 \\
Res50-GKT-TSA & 18.42 & \textbf{21.30} & 19.45 & 18.41 & 32.30 & \textbf{36.97} & 33.10 & 31.55 \\
Res50-GKT-RTF & 19.70 & \textbf{20.87} & 18.93 & 18.13 & 34.90 & \textbf{36.37} & 33.79 & 31.95 \\
VoV-SCA-TSA & 33.01 & 33.36 & \textbf{33.50} & 33.13 & 46.61 & 46.99 & \textbf{47.06} & 46.83 \\
VoV-SCA-RTF & 31.97 & 32.22 & \textbf{32.31} & 31.59 & 44.49 & 45.31 & \textbf{45.72} & 45.35 \\
VoV-GKT-TSA & 30.94 & 32.52 & \textbf{32.56} & 30.72 & 44.37 & 46.06 & \textbf{46.24} & 44.46 \\
VoV-GKT-RTF & 30.06 & 31.07 & \textbf{31.28} & 29.97 & 42.99 & 44.51 & \textbf{44.77} & 43.72 \\
\bottomrule
\bottomrule
\end{tabular}
\label{tab:ft10comp}
\end{table*}

\begin{table*}[!t]
\centering
\renewcommand{\arraystretch}{1.5}
\caption{Ablation study of the weight essemble method, we report performance on 70\% Nuscenes Validation Datset.}
\begin{tabular}{c|>{\centering\arraybackslash}p{1.0cm}>{\centering\arraybackslash}p{1.5cm}>{\centering\arraybackslash}p{1.5cm}>{\centering\arraybackslash}p{1.5cm}|>{\centering\arraybackslash}p{1.0cm}>{\centering\arraybackslash}p{1.5cm}>{\centering\arraybackslash}p{1.5cm}>{\centering\arraybackslash}p{1.5cm}}
\toprule
\toprule
\multirow{2}{*}{Model} & \multicolumn{4}{c|}{mAP  $\uparrow$ (\%)} & \multicolumn{4}{c}{NDS $\uparrow$ (\%)} \\
\cline{2-9}
 & Baseline & MML-Average & MML-Softmax & MML-Greedy & Baseline & MML-Average & MML-Softmax & MML-Greedy \\
\hline
Res50-SCA-TSA & 15.08 & \textbf{15.11} & 14.00 & 13.65 & 37.66 & \textbf{37.68} & 35.35 & 34.80 \\
Res50-SCA-RTF & 14.23 & \textbf{14.33} & 13.47 & 12.72 & 36.21 & \textbf{36.46} & 34.14 & 33.42 \\
Res50-GKT-TSA & 13.04 & \textbf{14.31} & 13.43 & 13.08 & 33.68 & \textbf{36.34} & 34.31 & 33.84 \\
Res50-GKT-RTF & 13.57 & \textbf{13.76} & 12.69 & 12.47 & 35.61 & \textbf{35.62} & 33.84 & 33.64 \\
VoV-SCA-TSA & 21.21 & \textbf{21.55} & 21.36 & 21.52 & 43.06 & \textbf{43.31} & 43.02 & \textbf{43.31} \\
VoV-SCA-RTF & 19.85 & 20.06 & \textbf{20.41} & 19.81 & 40.71 & 41.77 & \textbf{42.04} & 41.02 \\
VoV-GKT-TSA & 20.11 & \textbf{20.80} & 20.72 & 20.12 & 41.92 & 42.85 & \textbf{42.93} & 42.06 \\
VoV-GKT-RTF & 19.02 & \textbf{19.42} & 19.31 & 19.11 & 40.07 & \textbf{40.99} & 40.80 & 40.31 \\
\bottomrule 
\bottomrule
\end{tabular}
\label{tab:ft30comp}
\end{table*}

\subsection{Ablations}
In this section, we conduct the following ablation studies to verify the effectiveness of the proposed MML-Ave training strategy, leveraing a simple and efficient averaging operation. We further conduct numerical experiments with different model weight merge algorithms, as shown in Fig.\ref{fig:ABS}, termed MML-Softmax and MML-Greedy respectively. In MML-Softmax, module parameters are updated through a weighted summation, where the parameters are determined by factors derived from the accuracy metric and the softmax function as show in Eq.\eqref{eq:softmax}. MML-Greedy employs a greedy method and further prioritizes simplicity, where module weights are updated using the parameters belonging to the model exhibiting the highest mAP score.

\begin{equation}
\begin{aligned}
\mathbf{\omega(\theta_i)} = \frac{e^{\mathbf{Val\_mAP}(\theta_i)}}{\sum_{j=1}^K e^{\mathbf{Val\_mAP}(\theta_j)}}
\end{aligned}
\label{eq:softmax}
\end{equation}

\begin{figure}[htbp]
    \centering
    \includegraphics[width=1.0 \linewidth]{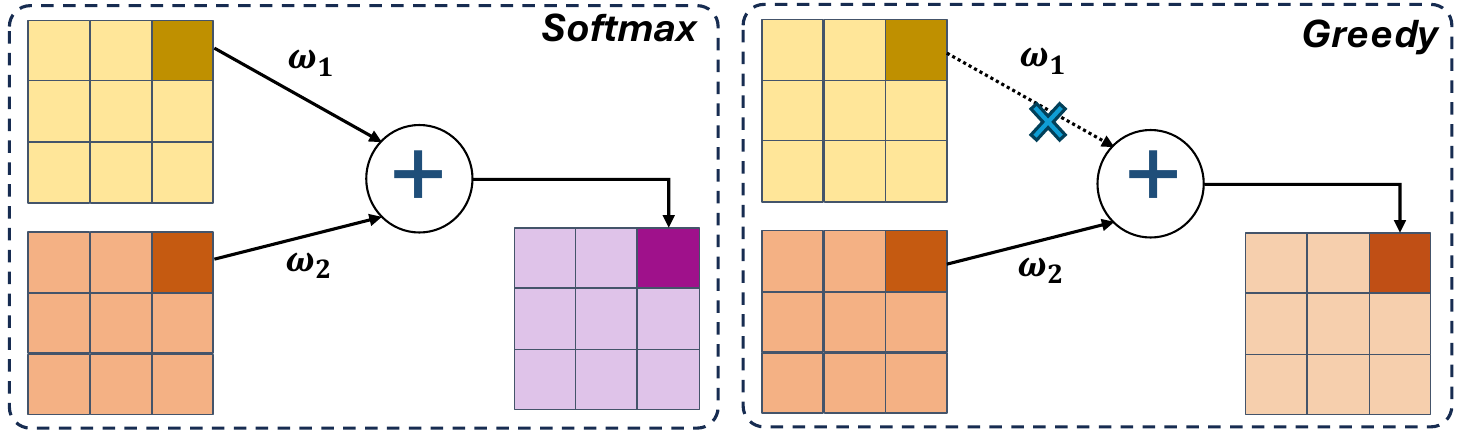}
    \caption{Illustration between (a) MML-Softmax and (b) MML-Greedy.}
\label{fig:ABS}
\end{figure}


We quantitatively compare our MML-Average method against MML-Softmax and MML-Greedy with respect to the nuScenes test split, as shown in Table~\ref{tab:ft10comp} and Table~\ref{tab:ft30comp}. The quantitative comparison results, under different data settings, underscore the performance of various ensemble strategies on ResNet50 and VoVNet models. As can be seen in the Table~\ref{tab:ft10comp}, the ResNet series models register a marginal yet significant mAP enhancement from 21.41\% to 22.29\% and an NDS improvement from 36.62\% to 37.98\% with the MML-Average strategy. In contrast, the MML-Softmax and MML-Greedy strategies exhibit reduced mAP and NDS values, highlighting their inferiority to the MML-Average approach. VoVNet series models, such as VoV-SCA-TSA, reveal a subtle but consistent mAP and NDS increase with the MML-ave strategy, achieving 33.50\% and 47.06\% respectively.

In order to further verify the effectiveness of the proposed MML-Average method, 30\% of the data are randomly selected from the validation set as training data, while the remaining 70\% of the data are used as test sets. The 30\% nuScenes dataset corroborates these results, with MML-Ave yielding a slight mAP and NDS uptick for ResNet50 models, whereas MML-Softmax and MML-Greedy fail to surpass the baseline. Notably, VoVNet series models manifest more pronounced improvements with MML-ave, as seen in the VoV-SCA-TSA model's mAP and NDS rise to 21.55\% and 43.31\%. The VoV-SCA-RTF model echoes this pattern, with mAP and NDS advancements to 20.06\% and 41.77\%.


\begin{figure*}[htbp]
	\centering
	\includegraphics[width=\textwidth]{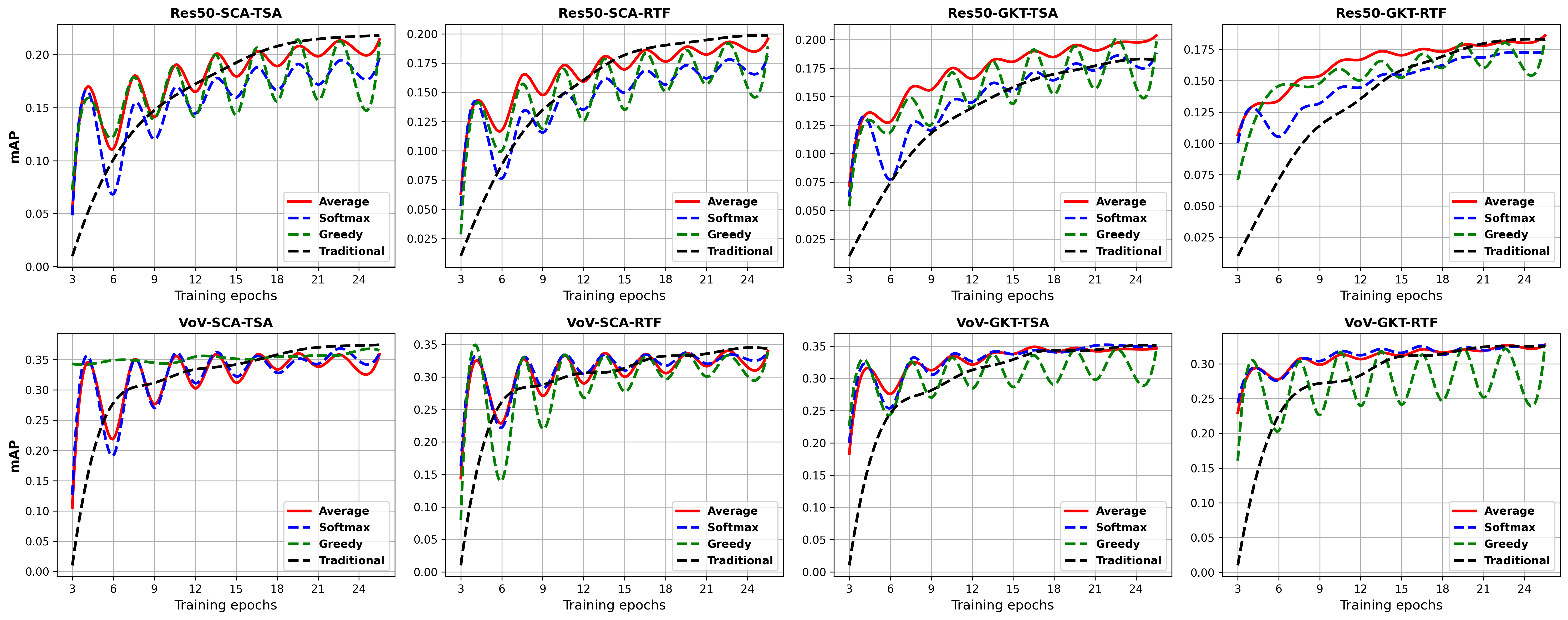}
        \caption{Observations about different weight merging strategies. Each subfigure displays the mAP indicator of each model over the course of pre-training.}
\label{fig:trend_8map}
\end{figure*}

\begin{figure*}[htbp]
	\centering
	\includegraphics[width=\textwidth]{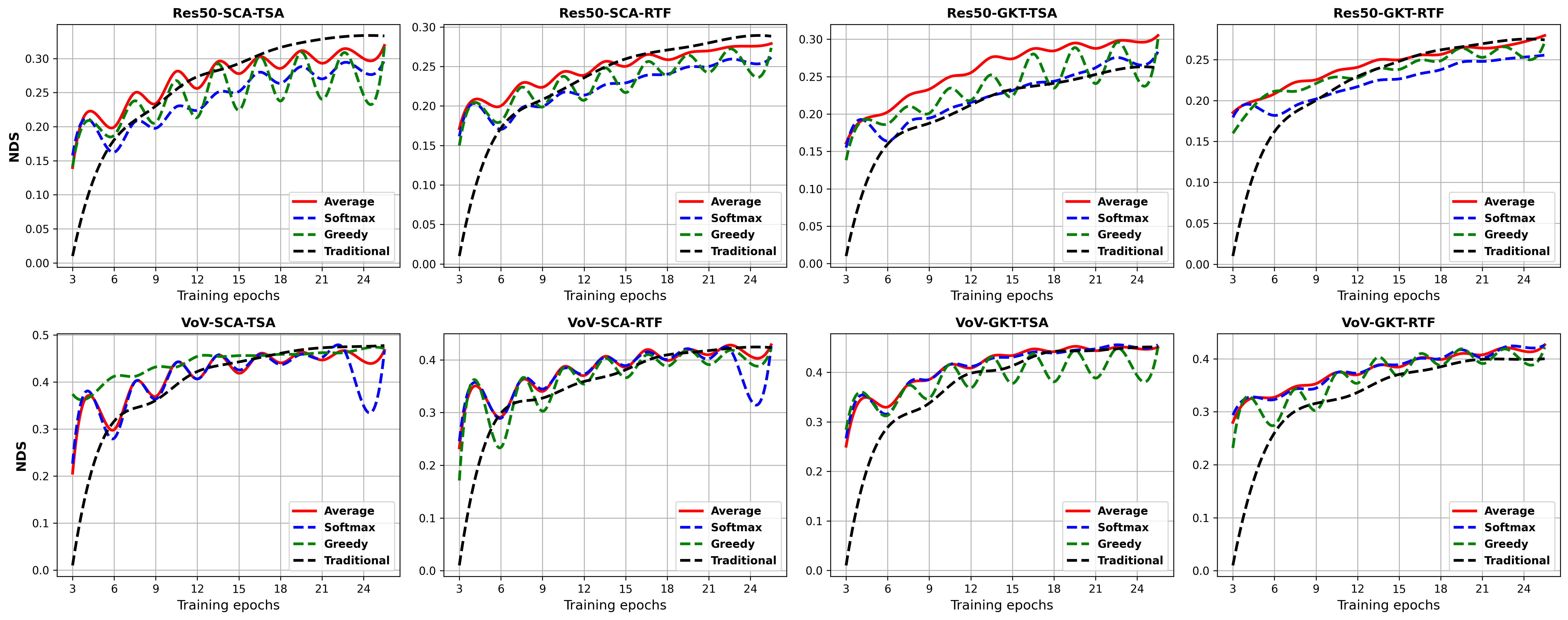}
        \caption{Observations about different weight merging strategies. Each subfigure displays the NDS indicator of each model over the course of pre-training.}
\label{fig:trend_8nds}
\end{figure*}

In general, the MML-ave strategy provides the most consistent improvements across both ResNet50 and VoVNet series models, enhancing both mAP and NDS metrics. Conversely, the MML-softmax and MML-greedy strategies often result in lower generalization performance compared to the baseline, suggesting that these optimization approaches may not be as effective for the models and dataset used in this study.


\subsection{Discussion}
In this section, we have witnessed the performance of all of the eight models in the pre-training stage. The fitting curves presented in Fig {\ref{fig:trend_8map} and \ref{fig:trend_8nds}} delineate the performance trends of ensemble models, showcasing the incremental improvements and stability of the MML-Ave, MML-Softmax, MML-Greedy and Traditional training approaches over time. The model performance is measured over a span of epochs, with the x-axis representing the progression through training epochs and the y-axis depicting mAP and NDS indicators, respectively. The subplots illustrate a distinct performance trajectory for each optimization strategy. The Traditional training method is consistently represented by a black curve, which demonstrates a gradual and steady increase in performance across all subplots. The MML-Ave and MML-Greedy based strategies, indicated by red and green curves respectively, exhibit competitive and often overlapping improvements, with the MML-Greedy strategy showing a slight edge in performance over the MML-Ave strategy in most cases. The MML-Softmax strategy, represented by a blue curve, consistently displays a more conservative rate of improvement compared to the other strategies, suggesting a potentially different convergence profile.

As can be seen in Fig {\ref{fig:trend_8map}, the MML-Ave approach shows a gradual increase in performance, reaching a plateau, indicating a consistent performance that stabilizes after a certain number of epochs. The MML-Softmax approach appears to perform slightly better than MML-Average, suggesting that the incorporation of weights in the algorithm might provide a modest improvement in performance. The MML-Greedy approach exhibits fluctuations, which could imply that it capitalizes on certain epochs more effectively, though it may also be more susceptible to overfitting or less generalizable. In contrast, traditional training approach maintains a lower performance level throughout the epochs, indicating a more stable but potentially less optimized performance compared to the other methods.

\begin{table*}[htbp]
\centering
\renewcommand{\arraystretch}{1.5}
\caption{Comparison results of all the 8 models on nuScenes val dataset in the pre-training stage.}
\begin{tabular}{c|>{\centering\arraybackslash}p{1.0cm}>{\centering\arraybackslash}p{1.5cm}>{\centering\arraybackslash}p{1.5cm}>{\centering\arraybackslash}p{1.5cm}|>{\centering\arraybackslash}p{1.0cm}>{\centering\arraybackslash}p{1.5cm}>{\centering\arraybackslash}p{1.5cm}>{\centering\arraybackslash}p{1.5cm}}
\toprule
\toprule
\multirow{2}{*}{Model} & \multicolumn{4}{c|}{mAP  $\uparrow$ (\%)} & \multicolumn{4}{c}{NDS $\uparrow$ (\%)} \\
\cline{2-9}
 & Baseline & MML-Average & MML-Softmax & MML-Greedy & Baseline & MML-Average & MML-Softmax & MML-Greedy \\
\hline
Res50-SCA-TSA & \textbf{21.82} & 21.44 & 19.73 & 21.21 & \textbf{33.31} & 31.91 & 29.78 & 31.68 \\
Res50-SCA-RTF & \textbf{19.83} & 19.58 & 18.10 & 18.93 & \textbf{28.83} & 27.90 & 26.16 & 27.35 \\
Res50-GKT-TSA & 18.21 & \textbf{20.38} & 18.85 & 19.86 & 26.16 & \textbf{30.50} & 28.32 & 30.01 \\
Res50-GKT-RTF & 18.31 & \textbf{18.62} & 17.33 & 18.26 & 27.41 & \textbf{27.95} & 25.55 & 27.15 \\
VoV-SCA-TSA & \textbf{37.46} & 35.88 & 36.11 & 36.54 & \textbf{47.73} & 46.75 & 46.69 & 47.11 \\
VoV-SCA-RTF & \textbf{34.32} & 34.02 & 33.88 & 33.87 & 42.32 & \textbf{42.85} & 42.60 & 42.17 \\
VoV-GKT-TSA & 35.06 & 34.67 & \textbf{35.09} & 34.59 & 45.08 & 45.08 & \textbf{45.55} & 45.14 \\
VoV-GKT-RTF & 32.51 & \textbf{32.73} & 32.69 & 32.67 & 40.00 & 41.33 & \textbf{41.48} & 41.45 \\
\bottomrule
\bottomrule
\end{tabular}
\label{tab:precomp}
\end{table*}

The experimental results from the nuScenes validation set, as presented in Table \ref{tab:precomp}, offering insights into the performance implications of different ensemble strategies. The comparison reveals that the MML-Ave and MML-Greedy strategies generally lead to diminished performance when compared to the Baseline across the ResNet50 models. For instance, the Res50-SCA-TSA model shows a decrement in NDS from 21.82\% (Baseline) to 19.73\% (Greedy) and a similar decline in mAP from 33.31\% to 29.78\%, respectively. In stark contrast, the VoVNet models, such as VoV-SCA-TSA and VoV-SCA-RTF, display a higher degree of consistency and stability in performance under the same strategies. The VoV-SCA-TSA model maintains a NDS above 35.88\% and an mAP above 46.69\%, even under the Greedy strategy, which suggests a more robust handling of strategy-induced variations.

The resilience of VoVNet models, particularly under the Average and Greedy strategies, is indicative of their potential superiority in multi-module ensemble configurations. The VoV-SCA-RTF model, for example, only experiences a slight fluctuation, with a NDS of 33.88\% and an mAP of 42.60\% under the Greedy strategy, compared to the Baseline scores of 34.32\% and 42.32\%, respectively. In summary, the VoVNet architecture appears to be more adept at sustaining performance across different strategies, whereas the ResNet50 models are more susceptible to strategy-induced performance variations. This robustness of VoVNet models may be attributed to their capticity of representation learning.

\section{Conclusion}


In this paper, Twe propose a novel modularized perception scheme, tailored to the next-generation automotive computing intelligent platform.  The proposed hierarchical and decoupled BEV learning framework provides a rich and flexible set of basic algorithmic building blocks, allowing developer to select and combine different functional modules according to specific requirements, thereby facilitating the rapid development of customized intelligent driving perception algorithms.  The framework enables continuous expansion of the functional module library, adapting to diverse user needs and featuring continuous learning capabilities, thereby reducing development cycles.

Our paper is a significant step towards establishing a new paradigm for the development of intelligent driving perception systems.  The hierarchical and decoupled scheme, the Multi-Module Learning method, and the empirical validation on nuScene dataset collectively contribute to a framework that is not only efficient and accurate but also adaptable and forward-looking.  We anticipate that our work will serve as a catalyst for further innovation in the field and inspire a new generation of researchers and practitioners to push the boundaries of what is possible in intelligent driving perception.


\bibliographystyle{IEEEtran}
\bibliography{reference}
\newpage

\end{document}